\documentclass[journal]{IEEEtran}

\ifCLASSINFOpdf
\else
   \usepackage[dvips]{graphicx}
\fi
\usepackage{url}

\usepackage{graphicx}
\usepackage{subcaption}
\usepackage[ruled,vlined]{algorithm2e}
\usepackage{bbm}
\usepackage{epstopdf}
\usepackage{amsfonts} 
\usepackage{color}
\usepackage{amssymb}
\usepackage[noadjust]{cite}
\usepackage{bm,nicefrac}
\usepackage{amstext}
\usepackage{ntheorem}
\usepackage{flushend}
\newtheorem{theorem}{Theorem}
\usepackage{mathtools}
\usepackage{algorithmic}
\usepackage[font=small]{caption} 

\newcommand{\cblue}[1] {{\color{black}{#1}}}

\newcommand{\bI}{{\bf I}}

\newcommand{\Real}{\mathbb{R}}
\def\x{{\mathbf x}}
\def\y{{\mathbf y}}

\def\p{{\mathbf p}}

\newcommand{\E}{\mathrm{E}}

\begin{document}

\title{Hamiltonian Adaptive Importance Sampling}

\author{Ali Mousavi, Reza Monsefi, and  V\'ictor Elvira, \IEEEmembership{Senior Member, IEEE}, 
\thanks{A. Mousavi and R. Monsefi are with Computer Department, Engineering Faculty, Ferdowsi University of Mashhad (FUM), Mashhad, Iran (e-mail: mousavi@iau-neyshabur.ac.ir; monsefi@um.ac.ir).}
\thanks{V. Elvira is with the School of Mathematics, University of Edinburgh, United Kingdom (e-mail: victor.elvira@ed.ac.uk).}}

\markboth{IEEE SIGNAL PROCESSING LETTERS, Vol. XX, No. X, XXXX}
{Shell \MakeLowercase{\textit{et al.}}: Bare Demo of IEEEtran.cls for IEEE Journals}
\maketitle

\begin{abstract}
Importance sampling (IS) is a powerful Monte Carlo (MC) methodology for approximating integrals, for instance in the context of Bayesian inference. 
In IS, the samples are simulated from the so-called proposal distribution, and the choice of this proposal is key for achieving a high performance.
In adaptive IS (AIS) methods, a set of proposals is iteratively improved.
AIS is a relevant and timely methodology although many limitations remain yet to be overcome, e.g., the curse of dimensionality in high-dimensional and multi-modal problems. 
%
%
Moreover, the Hamiltonian Monte Carlo (HMC) algorithm has become increasingly popular in machine learning and statistics.
 HMC has several appealing features such as its exploratory behavior, especially  in high-dimensional targets, when other methods suffer. 
 In this paper, we introduce the novel Hamiltonian adaptive importance sampling (HAIS) method. \cblue{HAIS implements a two-step adaptive process with parallel HMC chains that cooperate at each iteration. The proposed HAIS efficiently adapts a population of proposals, extracting the advantages of HMC. HAIS can be understood as a particular instance of the  generic layered AIS family with an additional resampling step.} 
 HAIS achieves a significant performance improvement in high-dimensional problems w.r.t. state-of-the-art algorithms. 
 We discuss the statistical properties of HAIS and show its high performance in two challenging examples.
\end{abstract}



\IEEEpeerreviewmaketitle

\section{Introduction}
\IEEEPARstart{I}{n} statistical signal processing, many tasks require the computation of expectations with respect to a probability density function (pdf).
 Unfortunately, obtaining closed-form solutions to these expectations is infeasible in many real-world challenging problems. 
 There are various approximation techniques to solve this problem, the most popular of which is the Monte Carlo (MC) methodology, based on the generation of random samples \cite{robert2monte}. 
 Arguably, the two main subfamilies of MC methods are importance sampling (IS) \cite{elvira2021advances} and Markov chain Monte Carlo (MCMC) \cite[Chapter 6]{robert2monte}.
 %
 %
 In this paper, we focus on the IS methods where the samples are obtained by simulating from the so-called proposal distribution.
The key of IS is an appropriate choice of the proposal distribution, which is a hard and very relevant problem. 
%
Since choosing a good proposal in advance is in general unfeasible, adaptive IS (AIS) methods adapt a mixture of proposals, iteratively improving the quality of the estimators by better fitting the proposals \cite{oh1992adaptive}. 
 %
 %
 There are several families of AIS methods, such as the population Monte Carlo (PMC) \cite{cappe2004population,cappe2008adaptive,elvira2017improving}, the AMIS algorithm \cite{cornuet2012adaptive,el2019efficient}, or gradient-based techniques \cite{elvira2015gradient,schuster2015gradient,fasiolo2018langevin}.  
\cblue{The research in AIS continues being very active and many crucial challenges remain open (see \cite{bugallo2017adaptive} for a recent survey).}
%
For instance, high-dimensional and multi-modal targets are particularly challenging to be explored and most AIS (and adaptive MCMC) methods fail to efficiently discover regions with relevant probability mass. 
%
%
Many efforts have been devoted to adapt the AIS proposals through an optimization process \cite{ryu2016convex,douc2007convergence, douc2007minimum,cappe2008adaptive,el2019stochastic}. 
Other works have addressed directly the reduction of the variability of the importance weight, ultimate responsible of the poor performance of the IS estimators  \cite{ionides2008truncated, koblents2015population, elvira2015efficient, elvira2016heretical,el2018robust,el2019recursive}. 
%
\cblue{Some recent methods aim at adapting the proposals through an independent process from the generated samples. We refer the interested reader to this class of AIS methods in \cite[Fig. 4(c)]{bugallo2017adaptive}. Particular instances of this class are the algorithms in the LAIS framework \cite{martino2017layered} or the techniques in \cite{rudolf2020metropolis,elvira2015gradient}. 
%
Unlike our proposed algorithm, the three explicit methods presented in the LAIS framework, which implement an adaptation of the upper layer based on the Metropolis-Hastings (MH) algorithm (see \cite{martino2017layered} for more details).
%
%
%
One limitation of these algorithms is the well-known random-walk behavior of the MH that makes the convergence of the  Markov chain  inefficient, especially in high-dimensional multi-modal distributions.
In addition, the performance of LAIS algorithms is highly dependent to the scale parameter of the proposals in the upper layer. 
}
Hamiltonian (or hybrid) Monte Carlo (HMC) \cite{duane1987hybrid, neal2011mcmc} is a state-of-the-art family of MCMC algorithms. 
%
\cblue{
HMC implements Hamiltonian dynamics allowing the samples to reach more distant points with higher probability of acceptance, which provides a greater improvement on exploratory capabilities compared with the other state-of-the-arts algorithms.
}
Despite the appealing properties of HMC,  the method can be notoriously difficult to tune (see \cite{mohamed2013adaptive,beskos2013optimal,mangoubi2017rapid}).  We note that HMC have been also incorporated to the SMC samplers framework \cite{buchholz2021adaptive}. 

\cblue{In this paper, we present the novel \emph{Hamiltonian adaptive importance sampling} (HAIS) algorithm which retains advantageous features from both HMC and AIS, achieving a high performance.
%
HAIS proposes a novel two-layered HMC-based AIS, inheriting the structure of LAIS or GAPIS. The upper layer consists of a two-step adaptation procedure that runs cooperative parallel HMC blocks in order to adapt the multiple proposals in AIS, which benefits both the exploratory behavior (particularly useful in multi-modal and high-dimensional problems) and the parallelization of the implementation. Another important contribution of the work is the consistent cooperation step where information among the chains is exchanged in order to enhance the global exploration.
\cblue{In addition, HAIS requires little tuning, overcoming one of the well-known limitations of HMC samplers.} 
Therefore, the validity of the estimators is ensured by \cblue{IS} arguments, unlike in HMC where the tuning can endanger the convergence of the Markov chain. 
%
}
The rest of the paper is organized as follows. In Section \ref{sec_problem},
we describe the problem while Section \ref{sec_HAIS} describes the novel HAIS method. Numerical examples are provided to compare \cblue{the} proposed HAIS with some other techniques in Section \ref{sec_Experiments}. 

\section{Problem statement}
\label{sec_problem}
\subsection{Bayesian inference}
\cblue{Let us consider a random variable of interest $ \x\in \mathbb{R}^{d_x} $ and $ \y \in \mathbb{R}^{d_y} $ be a set of related measurements or observations.} In the Bayesian framework, the variable of interest is characterized through the posterior probability function or the target pdf known as the 
\begin{equation}
	\widetilde{\pi}(\x|\y)=\frac{\ell(\y|\x)p_0(\x)}{Z(\y)},
\end{equation}
where \cblue{$ \ell(\y|\x) $ is the likelihood function}, $ p_0(\x) $, is the prior pdf, and $ Z(\y) $ is the model evidence (from now on we drop $\y$ in the notation). In many applications the goal is to obtain a moment of $ \x $ which can be expressed as the integral
\begin{equation}
	I = \E_{\widetilde{\pi}}[f(\x)]=\int_{\mathbb{R}^{d_{x}}} f(\x)\widetilde{\pi}(\x)d\x,
\end{equation}
where $ f: \mathbb{R}\rightarrow \mathbb{R}^{d_x} $ is some integrable function.

\subsection{Importance sampling}
Importance sampling (IS) is one of the main subfamilies of Monte Carlo methods. 
The basic idea of IS is to sample from a simpler pdf, the so-called proposal pdf $q(\x)$, to approximate the integrals w.r.t. to the \cblue{target} distribution $\widetilde\pi(\x)$ as
\begin{equation}
	 \hat{I}_{\text{IS}}(\x)=\frac{1}{\cblue{M}Z}\sum_{m=1}^{M}w_mf(\x_m),
	 \label{eq_uis}
\end{equation}
where $\left\lbrace\x_m\right\rbrace^{M}_{m=1}$ are iid samples generated from $q(\x)$, $Z$ is the normalization constant, and $w_m=\frac{\pi(\x_m)}{q(\x_m)}$ is the importance weight associated to $\x_n$. Then, we can construct $\hat{I}_{\text{IS}}$ which is an unbiased and consistent estimator of $I$, and its variance is related to discrepancy between $|f(\x)|\pi(\x)$ and $q(\x)$. \cblue{When $Z$ is unknown, the so-called \emph{self-normalized importance sampling} (SNIS) estimator can be constructed by plugging in Eq. \eqref{eq_uis} the unbiased estimator $\hat{Z}=\frac{1}{M}\sum_{m=1}^{M}w_m$ instead of $Z$ (see more details in \cite{elvira2019generalized}).} Since finding a good $q(\x)$ in advance is generally impossible, adaptive importance sampling (AIS) approaches are usually implemented, in order to iteratively improve the proposal. Relevant recent AIS methods are the PMC \cite{cappe2004population} and AMIS\cite{cornuet2012adaptive}, and more recently the LAIS \cite{martino2017layered}, DM-PMC \cite{elvira2017improving}, or \cite{elvira2019langevin} (see \cite{bugallo2017adaptive} for a review).

%

\subsection{Hamiltonian Monte Carlo}
\label{sec_HMC}
Hamiltonian Monte Carlo (HMC) is {a MCMC-based method that adopts Hamiltonian dynamics to explore the state space in order to propose future states in the Markov chain.} More precisely, let us denote $ U(\x)=-\log \pi(\x) $, which is usually called potential energy function in the physics literature \cite{duane1987hybrid,neal2011mcmc}. We consider also the kinetic energy function, $ V(\p)=\frac{1}{2}\p^T\bm{R}^{-1}\p $, with $ \bm{R} $ as a positive definite mass matrix and $ \p\in \Real^{d_x}$ as momentum vector. {The matrix $\bm{R}$ is typically diagonal or isotropic. Algorithm \ref{alg_hmc} describes the basic HMC procedure {which explores the joint probability density of $\x$ and $\p$, allowing for the simulation of samples.} Starting at an initial state $[\x_0,\p_0]$, \cblue{HMC} simulates Hamiltonian dynamics for $L$ steps using a discretization method. The common method is the leapfrog with the small step size parameter \cblue{$\epsilon$}. Next, the state of the position and momentum variables at the end of the simulation is used as the proposed state variables. Finally, $ (\x^*,\p^*) $ is accepted using an update rule analogous to the Metropolis acceptance criterion \cite{neal2011mcmc}.
By means of controlling the leapfrog size ($L$) and $\epsilon$, the acceptance rate of the HMC sampler can be adjusted \cite{leimkuhler2004simulating}. {The Hamiltonian dynamics benefit from several properties. 
Despite the significant benefits of HMC, especially in high dimension, HMC is known to be highly sensitive to the choice of parameters, particularly $ \epsilon $ and $ L $. \cblue{Choosing a too large step size will result in a low acceptance rate for new proposed state.} On the other hand, a too small step size will lead to slow exploration. Also, HMC encounters difficulties to sample from multi-modal distributions. 
We refer the interested reader  to \cite{neal2011mcmc,betancourt2017conceptual}.

\begin{algorithm}
\scriptsize
\SetAlgoLined
 \textbf{Input}: step size $ \epsilon $, leapfrog size $ L $, starting point $ \x^1 $, sample number $ M $ \\
 \For{$ m=1,...,M $}{
    Sample $ \p_0 \sim \mathcal{N}(0,\,1) $ and set $ \x_0=\x^m $ \\
    Run leapfrog method starting at $ (\x_0,\p_0) $ for $ L $ steps with step size $ \epsilon $  to obtain proposed states $ (\x^*,\p^*) $\\   
    Generate  $ u $ from $ U[0,1] $\\
    \textbf{if} {$ u\leq min[1,e^{U(\x^m)+V(\p^m)-U(\x^*)-V(\p^*)}] $}
    {   
    
     \hspace{5pt} $ \x^{m+1}=\x^* $
        
 	\textbf{else} 
 
     \hspace{5pt} $ \x^{m+1}=\x^m $
    }
 }
 \textbf{Output}: sample set $ \left\lbrace \x^m\right\rbrace_{m=1}^M $
 \caption{Hamiltonian Monte Carlo}
 \label{alg_hmc}
\end{algorithm}

\section{Hamiltonian Adaptive Importance Sampling}
\label{sec_HAIS}
In this section, we present the novel \emph{Hamiltonian adaptive importance sampling} (HAIS) method, which is summarized in Fig. \ref{fig:flowchart} and precisely described in Alg. \ref{alg_HAIS}.  HAIS is an iterative method that, at each iteration, performs three main operations: (a) sampling; (b) weighting; and (c) adaptation. 

\begin{figure}
    \centering
    \includegraphics[width=0.47\textwidth]{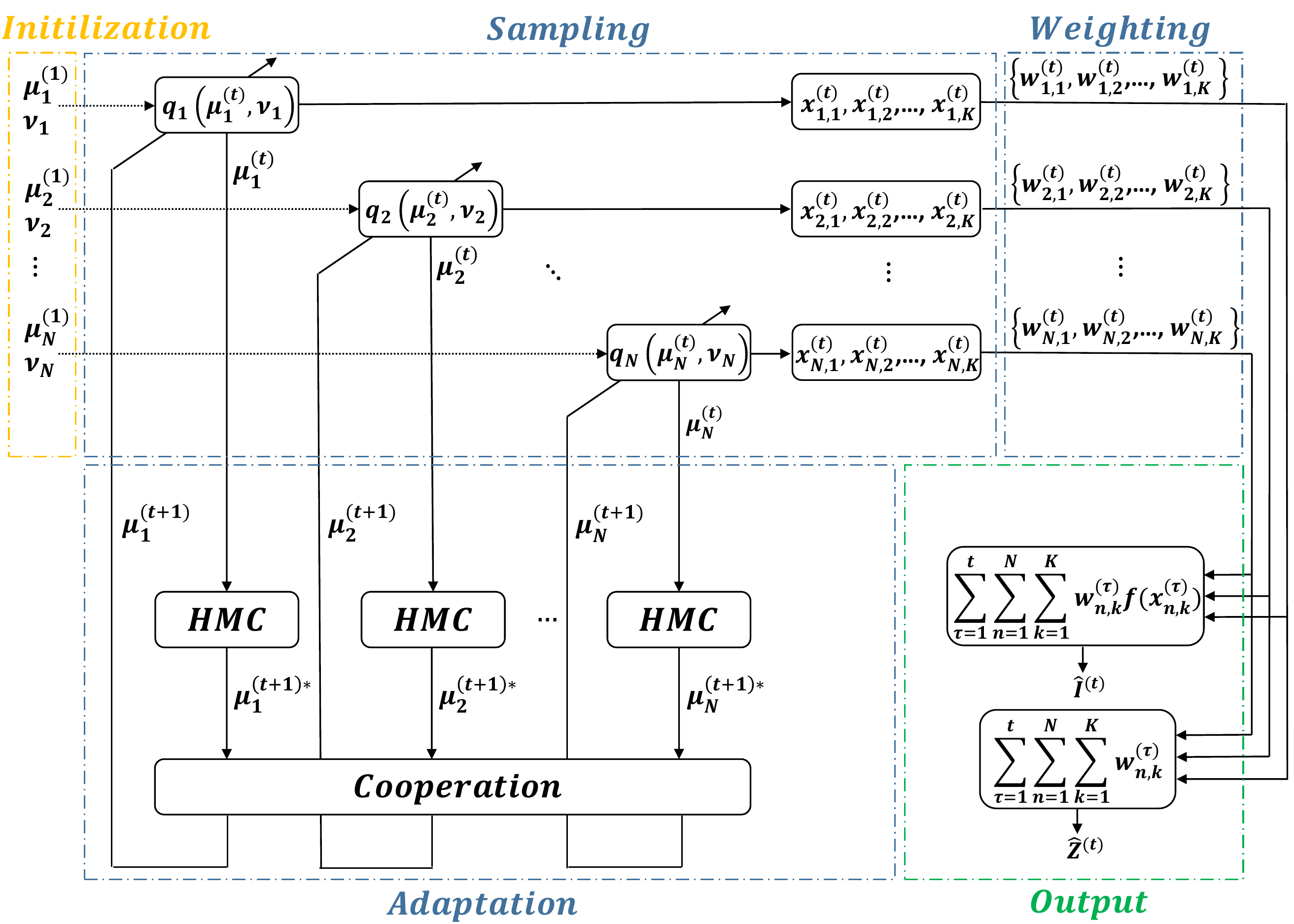}
    \caption{The flowchart of the proposed HAIS method showing the three main operations at iteration $t$ which results in the estimated values.}
    \label{fig:flowchart}
\end{figure}

\subsection{Algorithm description}
Algorithm \ref{alg_HAIS} describes the HAIS method. We consider $N$ (parametric)  proposal distributions, $ q_n(\x;\bm{\mu}_n,\bm{\nu}_n) $ where $ \bm{\mu}_n$ is the location parameter (e.g., the mean in a Gaussian pdf) and $\bm{\nu}_n$ contains the other parameters (e.g., $\bm{\nu}_n =\bm{\Sigma}_n$ is the covariance matrix in a Gaussian distribution). The algorithm starts with the initialization of the proposal parameters. At each iteration, $KN$ samples are generated from the $N$ proposals (exactly $K$ samples per proposal). Then, the importance weights are computed following deterministic mixture (DM) scheme. It has been recently shown that the unnormalized IS estimator of Eq. \eqref{eq_uis} with DM weights outperforms the same estimator if only the proposal $ q_n $ is in the denominator (instead of the whole mixture) \cite{elvira2019generalized}. Finally, adaptation procedure is performed. Next section is devoted to describe this adaptation.
%
\subsection{Adaptation process}
Fig. \ref{fig:flowchart} shows a two-step procedure in the adaptation of location parameters of the proposals (parallel HMC and cooperation).  
\cblue{
First, each HMC block explores the state space independently which is in a more efficient manner to discover local relevant features of the target, compared to other mechanisms such as MH-based methods or even naive gradient-based methods as in \cite{elvira2015gradient}.
The parallel structure amplifies the exploratory behavior, improving the local exploration capability of HAIS in high-dimensional multi-modal targets especially when the modes are distant.
}
Second, the information among the output of $ N $ parallel HMC is exchanged in order to improve the global exploration (see a discussion on local-global exploration in \cite{elvira2017improving}).
\cblue{
It is also important to remark that the cooperation step presents a theoretically sound and consistent procedure which does not require any free parameter to be tuned. We note that the adaptation of the proposal locations is completely independent from the samples for target estimation which helps the parallelization of the method (see a classification of AIS algorithm according to the adaptive mechanism in \cite{bugallo2017adaptive}).
}Note that, unlike HMC, the performance of HAIS does not critically depend on a precise tuning of the HMC parameters, since we use HMC for adapting the proposals not the final samples (that are properly weighted via IS).  
\cblue{
Assuming that all the HMC blocks obtain chains converging to the target pdf, the convergence of the overall HAIS is guaranteed.
} 
In the following we present a description of the novel approach. 

\subsubsection{\textbf{Parallel HMC \cblue{step}}}
We propose to run \cblue{$N$ independent HMC method} in a parallel way, each of which is shown as a HMC block in Fig. \ref{fig:flowchart}. Let us consider $ N $ parallel chains $ \{{\bm{\mu}}_n\}_{n=1}^N $ generated by those HMC blocks with $ \bm{\mu}_n^{(1)} $ as the initial d-dimensional starting point for the $ n $-th chain. We apply one iteration of $ N $ parallel chains, one for each $ \bm{\mu}_n^{(t)} $, returning $ \bm{\mu}_n^{(t+1)*} $ for $n=1,...,N$. Next, we compute the normalized DM weight of each $ \bm{\mu}_n^{(t+1)*} $,
\begin{equation}
	 \bar{w}_{\bm{\mu}_n^{(t+1)*}}\propto\frac{\pi(\bm{\mu}_n^{(t+1)*})}{\sum_{i=1}^{N}q_i(\bm{\mu}_i^{(t+1)*};\bm{\mu}_i^{(t)},\bm{\nu_i})}.
\end{equation}
Now we consider $ \widetilde{\pi}^{(t+1)*}(\bm{\mu}) $ as a random measure that approximates the target distribution, i.e.,
\begin{equation}
\label{eqn:randommeasure}
	\widetilde{\pi}^{(t+1)*}(\bm{\mu})=\sum_{n=1}^{N}\bar{w}_{\bm{\mu}_n^{(t+1)*}}\delta(\bm{\mu}-\bm{\mu}_n^{(t+1)*}),
\end{equation}
The theoretical motivation is that, after the burn-in periods, the $ N $ parallel HMC chains have converged to the target, so $ \bm{\mu}_n^{(t+1)*} \sim {\pi} $. 
As a result, the random measure based on weighted mean vectors in Eq. \eqref{eqn:randommeasure} approximates the target distribution.

\subsubsection{\textbf{Cooperation \cblue{step}}}
The final mean vector of the proposals for the next iteration are obtained by sampling from  $ \widetilde{\pi}^{(t+1)*}(\bm{\mu}) $, (via resampling), i.e., $ {\bm{\mu}}_n^{(t+1)} \sim \widetilde{\pi}^{(t+1)*}(\bm{\mu}) $. After resampling, a modified and unweighted random measure is produced as 
\begin{equation}
\label{eqn:randommeasure2}
		\widetilde{\pi}^{(t+1)}({\bm{\mu}})=\sum_{n=1}^{N}\delta({\bm{\mu}}-{\bm{\mu}}_n^{(t+1)}).
\end{equation}

In the following, we provide theoretical justification for using the cooperation step. The convergence of this adaptation process is given by Theorem \ref{theorem}.
\cblue{
\begin{theorem}
The moments of the random measure $ \widetilde{\pi}^{(t+1)}({\bm{\mu}}) $ obtained from outputs of the cooperation step (i.e., the means $ {\bm{\mu}}_n^{(t+1)} $) converge almost surely to those of the true distribution $\widetilde{\pi}(\x)$ when $N\to\infty$.
\label{theorem}
\end{theorem}
}
\textbf{Proof.} See the appendix.

\begin{algorithm}

\scriptsize
\SetAlgoLined
 \textbf{Input}: initial location, $ \{\bm{\mu}_n^{(1)}\}_{n=1}^N $, and scale, $ \{\bm{\nu}_n\}_{n=1}^N $, of proposals, $ N $ proposals, $ K $ sample per proposal, $ T $ iterations \\
 \For{$ t=1,...,T $}{
    (a) \textbf{sampling}: Draw K samples per individual proposal or mixand, 
    \[
             \x_{n,k}^{(t)} \sim q_n(\x;\bm{\mu}_n^{(t)},\bm{\nu}_n), n=1,...,N , k=1,...,K.
    \]
    (b) \textbf{weighting}: Compute the deterministic mixture (DM) weight of each sample,
    \[
            w_{n,k}^{(t)}=\frac{\pi(\x_{n,k}^{(t)})}{\sum_{i=1}^{N}q_i(\x_{i,k}^{(t)};\bm{\mu}_i^{(t)},\bm{\nu}_i)}.
    \]
    (c) \textbf{adaptation}: Update location of proposals for the next iteration through the \cblue{two-step adaptation procedure.}\\
    \hspace{5pt} 1. parallel HMC \cblue{step}: Generate new population of proposals, $ \{{\bm{\mu}}_1^{(t+1)*},{\bm{\mu}}_2^{(t+1)*},...,{\bm{\mu}}_N^{(t+1)*}\} $, by applying one iteration of $ N $ parallel HMC blocks.

    \hspace{5pt} 2. cooperation \cblue{step}: \cblue{Generate} final population of proposals, $ \{{\bm{\mu}}_1^{(t+1)},{\bm{\mu}}_2^{(t+1)},...,{\bm{\mu}}_N^{(t+1)}\} $, for the next iteration by sampling from equation (5).    
 }
 \textbf{Output}: a set of $KNT$ samples with associated weights, $ \{ \x_{n,k}^{(t)},w_{n,k}^{(t)} \}$ to obtain the estimation.
 \caption{Hamiltonian adaptive importance sampling}
 \label{alg_HAIS}
\end{algorithm}
\begin{table}

\caption{MSE in the approximation of the mean and normalizing constant of multi-modal target distribution.} 
\label{table}
\setlength{\tabcolsep}{5pt}
\centering
{\scriptsize
\begin{tabular}{|c||c|c||c|c||c|c||} \hline 
 {Method} & \multicolumn{2}{c}{$ \sigma=1 $} & \multicolumn{2}{c}{$ \sigma=2 $} & \multicolumn{2}{c||}{$ \sigma=5 $}    \\ \cline{2-7}
 &   $ E[\x] $ &   $ Z $    &  $ E[\x] $  & $ Z $      & $ E[\x] $   & $ Z $        \\ \hline 
GR-PMC 	& 64.92& 0.9418 & 64.14& 0.3126 & 65.70 & 0.3412   \\  
LR-PMC  & \textbf{21.57} & 0.9974 & 26.73 & 0.0998 & 65.24 & 0.9997    \\  
PI-MAIS ($ \lambda=5 $)  & 46.10 & 1 & 38.99 & 0.4531 & 36.63 & 0.5421  \\  
PI-MAIS ($ \lambda=10 $) & 60.51 & 1 & 60.30 & 1 & 53.28 & 0.9878   \\
 \hline
HAIS ($ \epsilon  = 5 $)      & 41.09 & \textbf{0.8649} & 17.57 & 0.0201 & 13.20 & 0.0031   \\ 
HAIS ($ \epsilon = 10 $)     & 42.76 & 0.8828 & \textbf{17.32} & \textbf{0.0162} & \textbf{12.87} & \textbf{0.0016}   \\ \hline 
\end{tabular}
}
\end{table}


\section{Numerical Examples}
\label{sec_Experiments}
\subsection{Bimodal target distribution in $ d_x=20 $}
In this section we consider a high-dimensional bi-modal target pdf in order to compare HAIS with some alternative methods. The target is a mixture of two Gaussians, i.e., $\pi(\x)=\sum_{i=1}^{2} \frac{1}{2}\mathcal{N}(\bm{\mu}_i,\,\bm{\Sigma}_i)$. Here $ \x\in \mathbb{R}^{20} $, $ \bm{\mu}_i=[{\mu}_{i,1},{\mu}_{i,2},...,{\mu}_{i,20}]^T $, and $ \bm{\Sigma}_i=c\bI_{20} $, for $ i\in \{1,2\} $ where $ \bI_{20} $ is an identity matrix of dimension  $ 20$. We set $ {\mu}_{1,d}=8 $ and $ {\mu}_{2,d}=-8 $ for all dimensions and also we set $ c=5 $. Multimodal settings are challenging, and in this example the two modes are distant, which over-complicates the exploration process at high dimension. Moreover, the proposal densities are Gaussian pdfs with uniformly selected initial means, i.e., $ \bm{\mu}_n \sim U([-4\times 4]^{d_x}) $ for $ n=1,...,N $, where none of the modes of the target fall within this area. We use the same isotropic covariance for all of proposals, $ \bm{\Sigma}= \sigma^2\bI_{20} $ with $ \sigma\in\{1,2,5\} $. We test for a constant number of samples, $ K=5 $, and number of proposals, $ N=100 $. For each algorithm the number of iterations, $ T $, is set so they have the same total number of target evaluations, $ E=2\times 10^5 $. We focus in estimation the mean and the normalizing constant of the target, which are obviously known in this example ($ \E[\x]=0 $ and $ Z=1$). We compared the proposed HAIS with recent high-performance AIS methods: GR-PMC and LR-PMC \cite{elvira2017improving}, and LAIS \cite{martino2017layered}. For the upper layer of the  LAIS,  we also consider Gaussian pdfs $ \varphi_n(\x|\bm{\mu}_n,\Lambda_n) $ where covariance matrices $ \Lambda_n=\lambda^2I_2 $ with $ \lambda \in \{5,10\} $. In our HAIS we consider all the HMC blocks to have the same step size parameter choosing from $ \epsilon \in\{5,10\} $ and a fixed value of $L=50$. 
The results are averaged over 200 independent runs. The simulation results are summarized in Table \ref{table} in terms of mean squared error (MSE) in the estimators. First, note that many settings/algorithms obtain very large MSE values. Those situations often correspond to the case where one or both modes are failed to be discovered. In the situation of missing both modes, the estimation of normalizing constant of the target is $ \hat{Z}\approx 0 $ which corresponds to a MSE$\approx 1$ as it happens in several settings. Second, we see that the HAIS algorithm generally outperforms the other methods. We note that the optimum value of $ \sigma $ which yields the smallest MSE, depends on the scale parameter of the target (here it is $ c=5 $).
\subsection{High-dimensional banana-shaped target distribution}
We now consider a benchmark multidimensional banana-shaped target distribution \cite{haario2001adaptive}, which is a challenging example because of its nonlinear nature. The target pdf is given by
\begin{align}\nonumber
	&\bar{\pi}(x_1,...,x_{d_x}) \\
	&\propto \exp\left(-\frac{x_1^2}{2\sigma^2}-\frac{\left(x_2+b(x_1^2-\sigma^2)\right)^2}{2\sigma^2}-\sum_{i=3}^{d_x}\frac{x_i^2}{2\sigma^2}\right).
\end{align}
We set $ b=3 $ and $ \sigma=1 $, then the true value for $ E_{\bar{\pi}}[\x]=0 $. Here Gaussian densities are considered as proposals with initial locations similar to the previous experiment. We use the isotropic covariance matrix for all of proposal pdfs, $ \bm{\Sigma}=\sigma^2\bI_{d_x} $ with $ \sigma^2=1 $. The selected values of $ K $, $ N $, and $ T $ is the same as in previous experiment. We compute the MSE in the estimation of $ E[\x] $ and the results are averaged over 200 Monte Carlo simulations. In order to compare the performance of the proposed method with other approaches as the dimension of the state space increases, we vary the dimension of the state space testing different values of $ d_x $ (with $ 2 \leq d_x \leq 50 $). Fig. \ref{fig:highdim} shows the log-MSE in the estimation of $ E[\x] $ as a function of the dimension $ d_x $ of the state-space, regarding the same techniques as in the previous bi-dimensional example. The performance of all the methods degrades as the dimension becomes larger. The result indicates that the novel HAIS scheme outperforms all the other methods under fair computational complexity comparison. 
\begin{figure}
  \centering
  \includegraphics[width=0.3\textwidth]{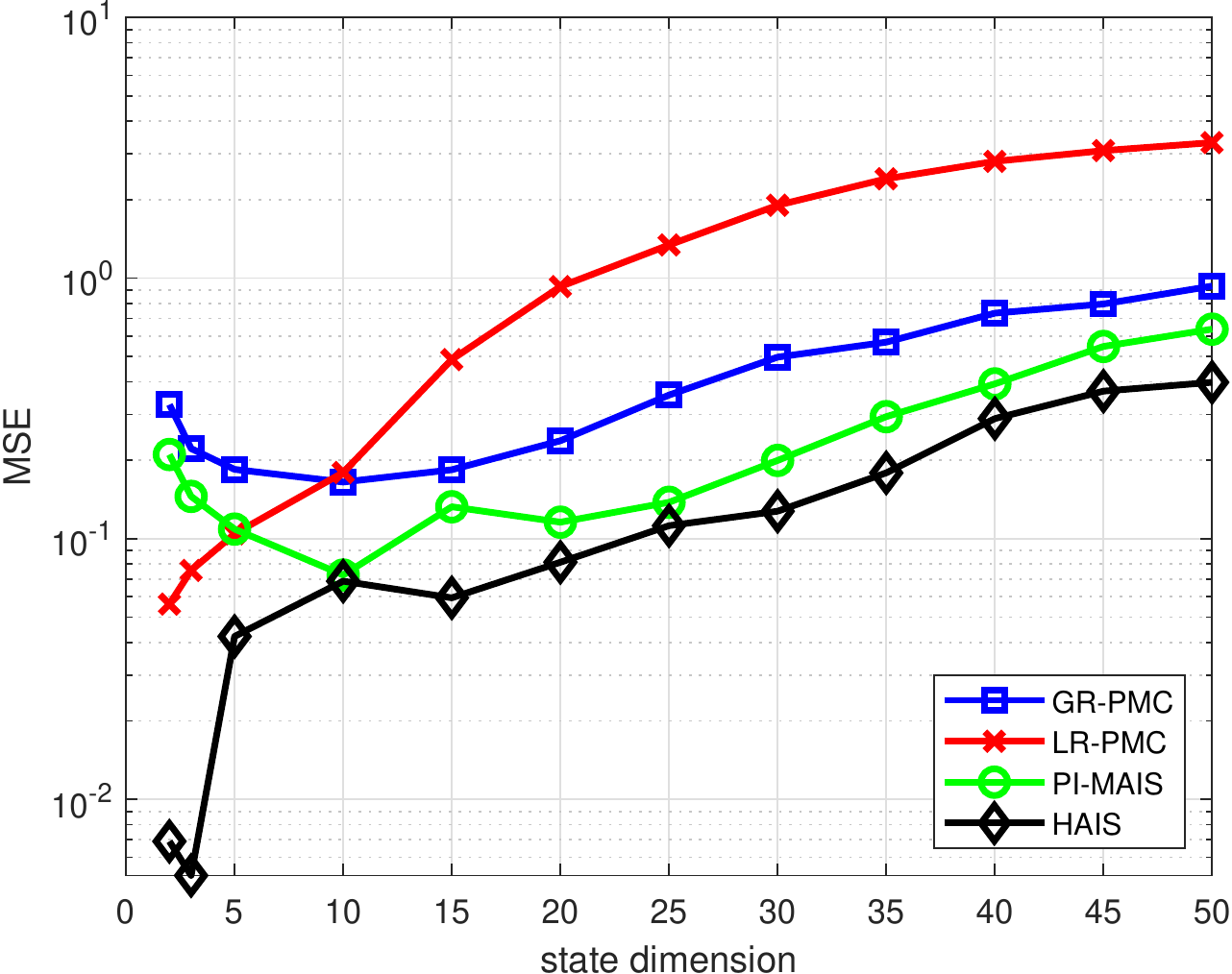}
\caption{log-MSE of $ E[\x] $, using $ N=100 $ proposals with $ \sigma=1 $, as the dimension of the target pdf, $ d_x $, increases.}
\label{fig:highdim}       
\end{figure}
\section{Conclusion}
In this paper we have proposed the HAIS algorithm, an adaptive importance sampler with particularly good behavior for high-dimensional and multi-modal problems. \cblue{HAIS belongs to the family of layered AIS samplers, and implements an adaptive procedure by running parallel HMC steps followed by a cooperation step, instead of the MH algorithm deployed in the algorithms of the LAIS framework}. The method is theoretically justified by HMC and resampling arguments. Simulation results have shown significant improvement compared with other state-of-the-art methods. 
\appendices
\section{Proof of Theorem 1}
\label{appendix}
The proof of Theorem 1 is based on a generalization of the result in \cite[Section 4.1]{elvira2017improving}. Let $ \bm{z}=[z_1,z_2,...,z_{d_x}] $ be a new auxiliary variable the same size as the variable of interest, $ \x=[x_1,x_2,...,x_{d_x}] $. Now consider the desired square integrable function to be an indicator function, i.e., $f_z(\x)=\prod_{d=1}^{d_x}\mathbbm{1}(x_d\leq z_d)$ where $ \mathbbm{1}$ denotes the indicator function. 
The integral $ I_z=\int f_z(\x)\pi(\bm{z})d\bm{z} $ becomes the multi-variate cumulative distribution function (cdf) of $ \pi(\x) $. In a similar way, we can obtain the cdf for $ \widetilde{\pi}^{(t+1)*}(\bm{\mu}) $, named $ \widetilde{I}_z $. It can be shown that $ \widetilde{I}_z \to I_z $ a.s. for any value of $z$ as $ N \to \infty $ \cite{geweke1989bayesian}. 
As a result, as $ N \to \infty $, the cdf associated to $ \widetilde{\pi}^{(t+1)*} $ a.s. converges to the target cdf. \cblue{Consequently, the outputs of the cooperation step are asymptotically distributed as the target $\widetilde\pi$, i.e., ${\bm{\mu}}_n^{(t+1)} \sim \pi(\x)$ when $N\to\infty$.} 

\bibliographystyle{ieeetr}

\end{document}